\documentclass[10pt,twocolumn,letterpaper]{article}

\usepackage[pagenumbers]{cvpr}      

\usepackage[table,xcdraw]{xcolor}

\usepackage{tikz}
\newcommand*\circled[1]{{\color{green}\tikz[baseline=(char.base)]{
            \node[shape=circle,draw,inner sep=1pt] (char) {#1};}}}
            
\definecolor{Gray}{gray}{0.9}
\renewcommand{\paragraph}[1]{\noindent\textbf{#1}\quad}

\newcommand{\vect}[1]{\boldsymbol{#1}}

\usepackage{graphicx}
\usepackage{booktabs}

\usepackage{amsmath,amsfonts,bm}

\def\eqref#1{equation~\ref{#1}}

\def\1{\bm{1}}

\def\vm{{\bm{m}}}

\def\vx{{\bm{x}}}

\def\vz{{\bm{z}}}

\DeclareMathAlphabet{\mathsfit}{\encodingdefault}{\sfdefault}{m}{sl}
\SetMathAlphabet{\mathsfit}{bold}{\encodingdefault}{\sfdefault}{bx}{n}

\usepackage{url}

\definecolor{cvprblue}{rgb}{0.21,0.49,0.74}
\usepackage[pagebackref,breaklinks,colorlinks,citecolor=cvprblue]{hyperref}

\def\paperID{} 
\def\confName{CVPR}
\def\confYear{2024}

\title{
Beyond Accuracy: Statistical Measures and Benchmark for Evaluation of Representation from Self-Supervised Learning\\
}

\author{
    Jiantao Wu$^1$\thanks{jiantao.wu@surrey.ac.uk}, Shentong Mo$^2$, Sara Atito$^1$, Josef Kittler$^1$, Zhenhua Feng$^1$, Muhammad Awais$^1$\\
$^1$ University of Surrey, $^2$ Carnegie Mellon University
}

\begin{document}

\twocolumn[{%
\renewcommand\twocolumn[1][]{#1}%
\maketitle
\begin{center}
    \centering
    \captionsetup{type=figure}
    \includegraphics[width=0.95\linewidth]{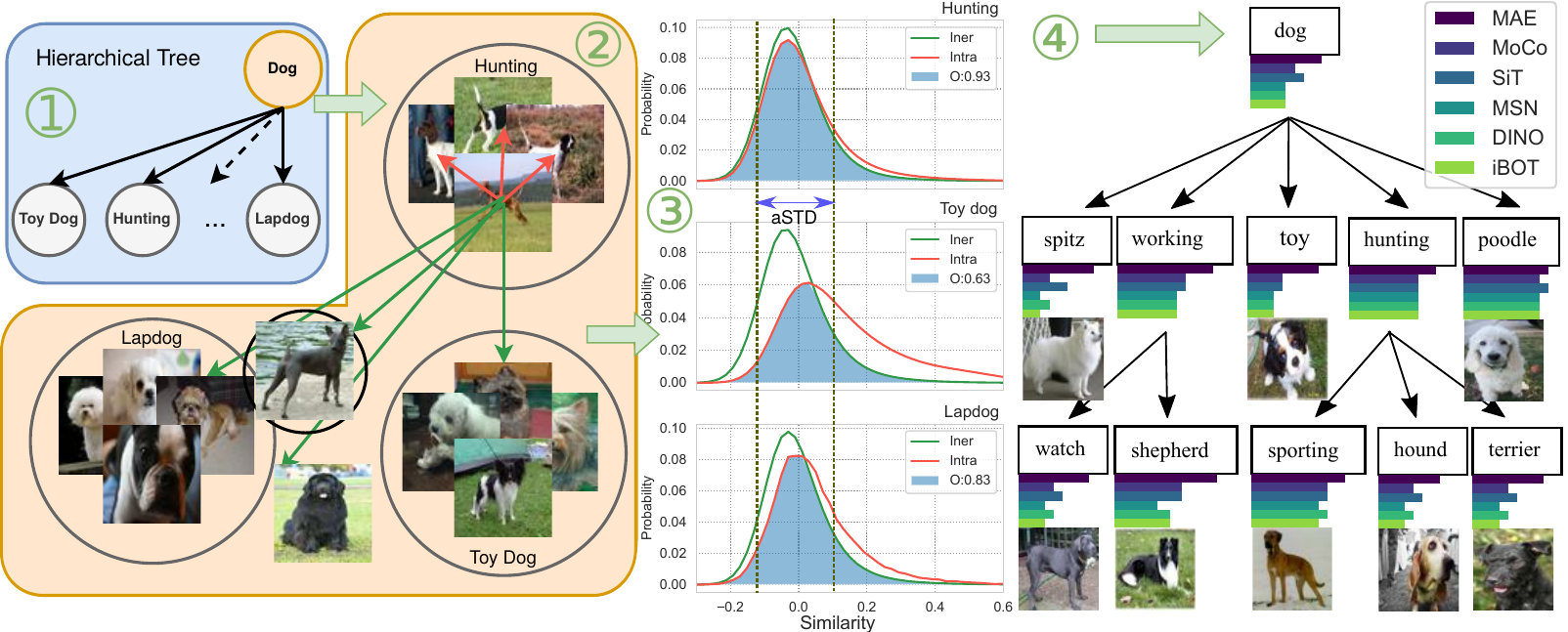}
    \vspace{-0.5em}
     \caption{Overview of evaluating synsets in WordNet by these steps: 1) Construct the hierarchical tree for a synset ``dog.n.01'', utilizing WordNet. 2) Build clusters for the direct children of the synset. Each cluster contains all samples belonging to the child in ImageNet-21K. 3) Compute both inter- and intra-similarity distributions for each cluster, and calculate the overlap (O) and average standard deviation (aSTD) to elucidate the degree of separability between clusters and the internal consistency of similarity distributions across clusters respectively. 4) Evaluate all synsets for SSL models and compare their overlap. The size of a color bar denotes the value of overlap for its corresponding model in the legend (lower the overlap better it is). }
    \label{fig:evaluation}
\end{center}%
}]

\begin{abstract}
Recently, self-supervised metric learning has raised attention for the potential to learn a generic distance function. It overcomes the limitations of conventional supervised one, e.g., scalability and label biases. Despite progress in this domain, current benchmarks, incorporating a narrow scope of classes, stop the nuanced evaluation of semantic representations. 
To bridge this gap, we introduce a large-scale benchmark with diversity and granularity of classes, Statistical Metric Learning Benchmark (SMLB) built upon ImageNet-21K and WordNet. SMLB is designed to rigorously evaluate the discriminative discernment and generalizability across more than 14M images, 20K classes, and 16K taxonomic nodes.  Alongside, we propose novel evaluation metrics -- `overlap' for separability and `aSTD' for consistency -- to measure distance statistical information, which are efficient and robust to the change of class number.
Our benchmark offers a novel perspective of evaluating the quality of representations beyond accuracy.
Our findings reveal the limitations of supervised learning and the class bias inherent in SSL models, offering insights into potential areas for future model enhancement. 
\end{abstract}

\section{Introduction}

Representation learning in computer vision has gained significant attention in recent years both in terms of supervised representation learning via metric learning~\citep{khalid2022npt, kaya_deep_2019,wang2015survey,yang2006distance} and self-supervised representation learning~\citep{chen_simple_2020,atito_sit_2021, atito_gmml_2022,he_masked_2022,oquab2023dinov2,mo2021spcl,wu2022objectwise,wu2023masked}. 
Metric learning aims to learn an optimal distance metric to measure the similarities of samples~\citep{kaya_deep_2019,wang2015survey,yang2006distance}, where samples from the same class will have small distances while samples from different classes will be far apart in the learnt embedding space. 
This capability of representation learning enables various tasks, notably in classification (e.g., facial recognition~\citep{DBLP:conf/icml/LiuWYY16}), clustering (e.g., image clustering~\citep{9830122,hoi2010semi}), and retrieval (e.g., image retrieval~\citep{DBLP:journals/corr/IRT}).
Typical metric learning methods like contrastive loss~\citep{hadsell_dimensionality_2006} and triplet loss~\citep{schroff_facenet_2015} require annotated data to form positive and negative pairs or triplets to optimize the distance metric.

However, supervised metric learning methods are designed for measuring the semantic distances in specific domains. For example, ArcFace~\citep{deng2019arcface}, trained on CASIA-WebFace~\citep{DBLP:journals/corr/YiLLL14a}, can only perform well on face-related tasks.
In addition, they have limitations on scaling to large datasets ~\citep{coghlan_good_2021,sun2023principle}. It relies on label information, which can be expensive and time-consuming to obtain. The finite class labels also can not capture the full semantic similarities between samples due to the complexity and uncertainty of real-world problems. In addition, the sampling of pairs or triplets can bias the embedding space because of label imbalance~\citep{rawat2021disentangling}.
All these obstruct learning a general distance metric to measure the semantic distance for arbitrary inputs.

Recent self-supervised learning methods address these limitations by pretraining on unlabeled data~\citep{chen_simple_2020,atito_sit_2021,atito_gmml_2022,caron_emerging_2021}. They design pretext tasks to learn semantic representations, which allows for learning more meaningful metrics without human annotations. For example, contrastive methods like SimCLR~\citep{chen_simple_2020} aim to pull together representations for different augmented views of the same image while pushing apart views from different images. 
Recently, self-supervised methods have shown great promise in learning representations with improved metrics over supervised counterparts~\citep{doersch_unsupervised_2015,kolesnikov_revisiting_2019,caron_emerging_2021,chen_simple_2020,atito_sit_2021,mo2022pauc,mo2023representation}. They are more data-efficient and can model fine-grained semantic similarities beyond discrete class labels~\citep{atito_sit_2021,caron_unsupervised_2020}. 
\citet{fu2021deep} proposed a self-supervised ranking framework to capture intra-class characteristics. 
Despite the use of SSL, it only plays as an auxiliary regularizer and still requires labels to learn discriminative features for inter-class samples.
\citet{Wang_Zeng_Chen_Dai_Xia_2022} gets rid of labels by contrastive quantization with code memory to learn binary hash codes for a dataset. However, they did not explore contrastive learning as a general distance measurement. A statistical framework of self-supervised metric learning in the context of multi-view data is developed by \citet{wang2022self}, yet, lacks a reality check for the existing family of SSL methods and on real problems. 
To bridge the gap, we conduct a large-scale empirical study on the evaluating semantic distinctions of learned representations from diversity and granularities.

The first obstacle is the lack of diversity and granularity of classes in the existing benchmarks. 
Current benchmarks specialize a narrow scope, such as Oxford and Paris Buildings~\citep{DBLP:conf/cvpr/RadenovicITAC18}, face identification~\cite{khalid2022npt} and Pets~\citep{parkhi2012cats}. Moreover, the flattened label structure in these benchmarks is infeasible to reveal granularity of classes. For instance, two dogs of subordinate breeds, e.g., Pekingese and Maltese, should be closer than two arbitrary dogs. 
Therefore, we do not know the how well the semantic representation is as a general semantic distance function to measure the similarity between two arbitrary images.
The second obstacle is the insufficiency of metrics for representations. The popular metrics, e.g., accuracy, cannot provide a complete picture of embedding space~\citep{musgrave_metric_2020}. These metrics, losing the distance information, only reveal a final task performance. 
We consider two desirable properties of representation \emph{separability} and \emph{consistency}, i.e., the degree to which distinct clusters are distinguishable from one another in the feature space, and the uniformity and reliability of similarity distributions across clusters.

To tackle the above problems, we build a novel benchmark, Statistical Metric Learning Benchmark (SMLB), on the top of ImageNet-21K~\citep{russakovsky2015imagenet} and WordNet~\citep{fellbaum1998wordnet}. This large-scale benchmark consists of 14,191,291 images divided into 20,498 classes and 16,632 taxonomic nodes.
This large-scale benchmark allows a wide examination of the model's ability to identify and differentiate nuanced distinctions within data, termed \emph{discriminative discernment}, as well as generalizability. 
We further propose a novel evaluation to measure the distance statistical information from the distributions between inter- and intra- pairs. To depict the characters of similarity distributions, we propose two novel metrics: \emph{overlap} to measure separability and \emph{aSTD} to measure consistency.
Figure~\ref{fig:evaluation} shows the overview of our work.
We create clusters for a synset based on its hierarchical tree and evaluate the corresponding overlap and aSTD. 
Then, we examine a wide range of SSL models with mono- or hyper- SSL techniques, e.g., view contrast, masked image modeling, and knowledge distillation on SMLB with \emph{overlap} and \emph{aSTD} to conduct an empirical study of learning semantic representations through SSL.
Finally, our study leads to important findings.
Our contributions can be summarized as:
\begin{itemize}
    \item We propose two metrics based on statistical information to evaluate the distance functions: \emph{overlap}  reveals separability, \emph{aSTD} reveals consistency.
    \item We build a large-scale Statistical Metric Learning Benchmark, \emph{SMLB}, to make a comprehensive evaluation of discriminative discernment and generalizability.
    \item From a bunch of experiments, we empirically find  1) the detrimental potential of supervised learning. 2) class bias of self-supervised learning.
\end{itemize}

\section{Related Work}

\paragraph{Benchmarks for Metric Learning.}
In metric learning, benchmarks are pivotal for evaluating the ability of algorithms to learn distance or representation functions for narrow tasks. For example, WebFace~\citep{DBLP:journals/corr/YiLLL14a} contains 494,414 face images of 10,575 real identities, which is used for face verification and face identification. Oxford and Paris Buildings~\citep{DBLP:conf/cvpr/RadenovicITAC18} are popular for image retrieval, which has 5,062 and 6,412 images from  11 landmarks. These benchmarks lack diversity to assess the representation for a general distance function to measure the semantic distance.
Task Adaptation Benchmark (VTAB)~\citep{DBLP:journals/corr/abs-2303-13505} is a benchmark with diverse, unseen tasks with few examples for a unified evaluation for general visual representation, yet, not designed for a distance measurement.
Though ImageNet-21K (IN21K)~\citep{russakovsky2015imagenet} was proven suitable for pretraining, there is no study to evaluate models, utilizing its 21K classes with hierarchical semantics from WordNet~\cite{fellbaum1998wordnet}.
In this work, we build a large-scale benchmark based on IN21K and WordNet for metric learning to validate discriminative discernment which refers to the capacity for making or recognizing detailed distinctions of the learned representations of SSL models.

\paragraph{Performance Measures for Metric Learning.}
Accuracy, Precision, Recall, F1-score, Normalized Mutual Information (NMI), and mean average precision (mAP) are popular metrics to report the performance.
\citet{musgrave_metric_2020} shows the weakness of accuracy metrics and concludes that it cannot provide a complete picture of embedding space, i.e., accuracy fails to show the separation of different classes. 
In addition, the accuracy loses distance information causing a lack of comprehensive assessment of representations.
Some studies highlight the importance of distance-based metrics for the representations.
Inter-class Similarity and Intra-class Deviation are two important aspects to promote the quality of representations~\citep{venkataramanan_tackling_2021}. 
\citet{DBLP:journals/corr/RippelPDB15} promote separability by penalizing class distribution overlap. 
In this work, we preserve the distance information by histograming similarity distributions and propose two metrics overlap and aSTD to evaluate their separability and consistency.

\paragraph{Evaluation Protocol.}
The finetune protocol is one of the most pivotal evaluations for SSL models~\citep{devlin_bert_2019}. Linear prob is another popular evaluation protocol, which does not change the representations and only trains a linear classifier. However, the above two protocols require lots of computational resources to train. What's worse, they change or re-weight the features to fit new tasks, cannot represent the encoding of meaningful information where the similarity in feature space corresponds to semantic distance.
Other prevalent evaluations, e.g., KNN and retrieval, are sensitive to the number of classes. This implies that these evaluations do not adjust or react in proportion to the complexity added by having more classes. For instance, 50\% accuracy in a binary classification task is not better than 40\% accuracy in a triple classification task.
All these motivate us to propose a flexible and unified evaluation protocol to evaluate semantic representations from these considerations.

\section{Proposed Evaluation Method}

\subsection{A Distance Perspective}

The main goal of this paper is to evaluate the SSL models from the distance perspective. 
A theoretical analysis is done by \citet{wang2022self} where they assume observed data samples, denoted as $X$, emanate from a confluence of latent factors and stochastic noise.The latent factor model draw the multi-view data $(\vect{X}_1,\dots, \vect{X}_n)$ as $\vect{X} = \vect{B} \vect{Z} + e$, where $\vect{Z}$ signifies the latent factors, $\vect{B}$ is a transformation matrix such that $\vect{B}^T\vect{B}=\Lambda $, and $e$ represents normally distributed noise. However, their analysis is designed for Mahalanobis distance, lacking a reality check for deep neural networks. 

To bridge the gap, we introduce an assumption on the modeling of factors for a class modulated by external labels, $Y$, that is, the appearance of attribution variable $Z_i$ is subjected to a Bernoulli distribution such that $Z_i|Y \sim \text{Bernoulli}(p_i)$.  For instance, the visibility of eyes in a face image could be influenced by whether the person is wearing sunglasses or not. Consequently, the probability of the factor’s (eye’s) presence or absence in the image of a human face. 

The distance, $s(\vect{X}_1, \vect{X}_2)= \sum_i \vect{Z}_1 \vect{Z}_2$, is ingeniously crafted as the dot product of the latent factor vectors, signifying the number of shared factors between two samples. Since the presence of each factor is determined probabilistically, the distance metric obeys a Poisson distribution or a normal distribution when there are enough attributions.
Therefore, the distance between classes reveals the count of shared attributions. We hypothesize that the distance between samples obeys a normal distribution and the pairs from the same class are significant closer than those from different classes.
In the following, we define \emph{intra} pairs as the pairs having the same class, and \emph{inter} pairs as the pairs having different classes. It is notable that the definition is flexible, depending on the content we are studying. The positive pairs can be two images with the same class or two patches of an object from one image. For simplicity, we adopt the former definition by default.

\paragraph{Similarity Definition.}
To compare the global features for images, we define the \emph{intra-class} pair as the pair of images from the same class, and \emph{inter-class} pair as the pair of images from different classes. For a given set of images, denoted as $\{\vx^i\}$, and their corresponding labels $\{y^i\}$, we can formulate the intra- and inter-class similarities as follows:
\begin{equation}
\begin{aligned}
    s^+_{y^i}(\vx^i,\vx^j) &= \mathit{sim}(\vz^i,\vz^j), \text{if}\ y^i = y^j, \\
    s^-_{y^i}(\vx^i,\vx^j) &= \mathit{sim}(\vz^i,\vz^j), \text{if}\ y^i \neq y^j.
\end{aligned}
\label{eq:class_similarity}
\end{equation}
To measure the similarity between features, we employ the cosine similarity function, denoted as $\mathit{sim}(\cdot,\cdot)$, which ensures the range of distance from -1 to 1.

\paragraph{Similarity Distribution.} 
The similarity distributions describe the likelihood of similarity for the distance of positive or negative pairs, which provides succinct summaries of distance information beyond accuracy. 
To calculate the probability density function (PDF) of similarity scores $s$ in condition, we have
\begin{equation}
    p(s|y) = \int_{\mathcal{Z}}\int_{\mathcal{Z}} p(s = \mathit{sim}(z1, z2) | y) \, p(z1, z2 | y) \, dz1 \, dz2
\end{equation}
where $p(z1, z2 | y)$ is the joint probability density of the feature vectors given the class label $y$, $\mathcal{Z}$ is the feature space for positive features or negative features. 
However, the real similarity distributions $p(s|y)$ are infeasible. We apply random sampling to get sample pairs and utilize a histogram from -1 to 1 with 100 bins to estimate their distributions. For each class, we can draw inter- or intra- similarity distributions $p^+(s|y), p^-(s|y)$. For example, we obtain the distributions by positive pairs for intra-similarity, denoted by red, and negative pairs for inter-similarity, denoted by green, as shown in Figure~\ref{fig:evaluation} \circled{2}.

\subsection{Evaluated Models}
We address the challenge of learning semantic distance measurement through the self-supervised pretraining of Vision Transformers (ViTs). Our problem setup is described using the following notation: An image, represented by $\vx$, is divided into non-overlapping patches $\vx_p$, with $p$ ranging from 1 to $N$, where $N$ is the total number of patches. A learnable classification (CLS) token, initialized with random weights, is appended to these patches. The ensemble of the CLS token and image patches is then processed by a transformer encoder, denoted as $f(\cdot)$, which concurrently extracts local and global features. The CLS token, $\vz=f(\vx)$, represents the entire image.

In the realm of self-supervised methods, three prominent and fundamental approaches are contrastive learning (CL)~\citep{assran2022masked_MSN,he_momentum_2020}, masked image modeling (MIM)~\citep{he_masked_2022,atito_gmml_2022}, self-distillation (SD)~\citep{zhou_ibot_2022,caron_emerging_2021}.
For mono-SSL, we examine three representative methods:
\textbf{MoCov3~\citep{chen_empirical_2021}} advances CL by using a momentum encoder to improve the consistency and quality of visual representations learned from different views of the same image. 
\textbf{MAE}~\citep{he_masked_2022} learns rich image representations by MIM to reconstruct the original image content from partially masked inputs. 
\textbf{DINO}~\citep{caron_emerging_2021} employs an SD approach, where a student network learns to predict the output of a momentum teacher network across different augmentations of an image.
For hybrid-SSL, we consider three models:
\textbf{iBOT}~\citep{zhou_ibot_2022} extends the SD framework of DINO by applying MIM to patch tokens.
\textbf{MSN}~\citep{assran2022masked_MSN} merges MIM with SD to cultivate detailed and robust representations.
\textbf{SiT}~\citep{atito_sit_2021} implements MIM on patch tokens and CL on the CLS token, enhancing the quality of representations in a synergistic manner.

We observe that feature normalization plays a vital role in accurately gauging semantic distances. Consequently, we standardize the features using the mean and standard deviation derived from the target dataset’s features. For further details, please refer to Section~\ref{sec:norm}.

\subsection{Proposed Metrics}
We propose two novel metrics, \emph{overlap} and \emph{average Standard Deviation (aSTD)}, to delve deeper into the nuances of similarity distributions and thereby gain a more insightful understanding of the features. These metrics are applied to a labeled dataset to identify positive and negative pairings. Unlike the conventional metric of \emph{accuracy}, which evaluates the end-to-end performance of a task, our proposed metrics focus on the statistical factors of features, specifically examining \emph{separability} and \emph{consistency}. We illustrate our metrics in Figure~\ref{fig:evaluation}~\circled{3}.

\paragraph{Overlap.}
\emph{Separability} indicates distinctiveness of features, meaning that features from different classes can be delineated by a decision boundary. In our analysis, we aim for the inter-class and intra-class distributions to be sufficiently distinct, allowing for a threshold that separates them. However, finding an optimal, consistent threshold for each class is challenging. As an alternative, we measure \emph{separability} using \emph{overlap}, which is the intersection area between the two distributions. The overlap is calculated using the probability distributions of distances or similarities within the same class (intra-class) and between different classes (inter-class), as shown in the following equation:
\begin{equation}
    O = \frac{1}{N} \sum_{y=1}^N \sum_{b\in \mathcal{B}} \min(p^{+}(s|y \in b) , p^{-}(s|y\in b)),   
\end{equation}
where $b$ is the bin of similarity range, $p^+(\cdot),p^-(\cdot)$ denote the probability of the similarity in the bin for positive and negative pairs, $N$ denotes the number of classes, $\mathcal{B}$ is the set of bin edges.

The overlap metric ranges from 0 to 1, reflecting the binary classification error when determining if a pair of features belongs to the same class. In an ideal scenario with only two classes, the classification performance is equivalent to finding an optimal similarity threshold for binary classification, with the goal of achieving zero overlap to ensure complete separability of inter-class and intra-class features.

\paragraph{Average Standard Deviation.}
\emph{Consistency} pertains to the stability of the distributions across all classes. We aim for the inter-class and intra-class distributions to maintain a consistent shape for each class. To quantify this, we introduce \emph{aSTD}, the average deviation of distributions across different classes, to capture the deviation within the dataset. This metric is particularly useful for assessing the diversity of a dataset with numerous classes that may vary significantly or be closely related semantically.
To calculate aSTD, we first determine the conditional similarity distributions ${p}(s|y)$ for each class label. We get the distribution by $p(s) = \frac{1}{N} \sum_{y=1}^N p(s|y)$. The average Standard Deviation is measured using the following formula:
\begin{equation}    
\begin{aligned}
\mu &= \sum_{b \in \mathcal{B}} p(s \in b) \cdot \mathrm{MEAN}(b), \\
\sigma &= \sqrt{\sum_{b \in \mathcal{B}} p(s \in b) \cdot ( \mathrm{MEAN}(b)- \mu)^2},
\end{aligned}
    \label{eq:std}
\end{equation}
where MEAN calculate the center of an edge, $ p(s \in b)$ denotes the probability of similarity falls in $b$.

aSTD serves as an indicator of the disparity among similarity distributions. A higher aSTD value suggests a greater degree of variance between these distributions, which implies that more effort is required to reconcile these differences. In practice, the aSTD of inter-similarity distributions across classes indicates the potential for performance enhancement through learning processes.

\subsection{Metric Validation}
In this section, we aim to substantiate the reliability and significance of our metrics. To this end, we trained MoCo v3, DINO, and MAE using ViT-S/16 on the IN1K dataset for 300, 200, and 800 epochs, respectively, and saved intermediary weights to obtain 10 checkpoints for each model. These models yield representations of varying quality, providing a broad spectrum of data points to demonstrate the correlation.

\begin{figure*}[tb]
    \centering
    \includegraphics[width=.85\linewidth]{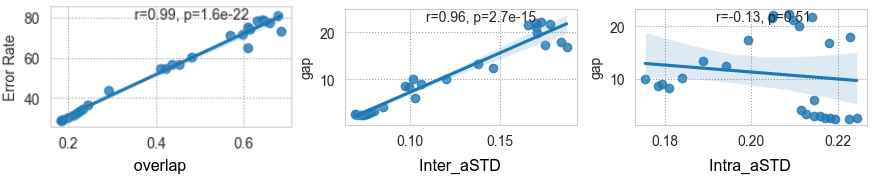}
    \caption{Overlap signifies separability and aSTD denotes consistency. ``error rate'' represents the performance on IN1K using k=10 for KNN. ``gap''  refers to the accuracy differential between linear probing and KNN. ``r'' and ``p'' denote Pearson product-moment correlation coefficient and the p-value associated for the two-sided hypothesis test, respectively.}
    \label{fig:overlap_acc}
\end{figure*}

To affirm the efficacy of overlap and aSTD, we assessed these models and compared our findings with standard evaluations. We utilized KNN (k=10) and linear probing (over 100 epochs) to measure accuracy on IN1K. Figure~\ref{fig:overlap_acc} illustrates the correlation between our metrics and both the error rate and the performance gap. The data reveal that overlap and inter-class aSTD correlate strongly with the error rate and gap (r=0.99, 0.96, respectively). In contrast, intra-class aSTD is not as important as inter-one. For convenience, we report aSTD denoting inter aSTD by default.  In conclusion, the overlap metric can serve as an indicator of the error rate associated with KNN performance, while aSTD signifies the precision of that indicator.

\begin{table*}[t]
\centering
\caption{Comparative analysis of overlap and aSTD for SSL models and DeiT on 16K synsets grouped by node levels. `AVG' denotes the average metrics across all nodes. MIM shows limited effectiveness in learning semantic representations. SD methods are much effective in extracting coarse semantics for low levels (1-9).}\label{tab:benchmark}
\setlength\tabcolsep{4pt} 
\begin{tabular}{lccccccccccccccc}
\toprule
      & \multicolumn{7}{c}{overlap $\downarrow$}                                  &  & \multicolumn{7}{c}{aSTD $\downarrow$}                                     \\
      \cmidrule(lr){2-8} \cmidrule(lr){10-16}
level & MAE     & MoCo  & SiT   & MSN   & DINO  & iBOT  & DeiT  &  & MAE   & MoCo  & SiT   & MSN   & DINO  & iBOT  & DeiT  \\
\hline
1   & \cellcolor[HTML]{959595}0.753 & \cellcolor[HTML]{ACACAC}0.696 & \cellcolor[HTML]{BBBBBB}\textbf{0.661} & \cellcolor[HTML]{A3A3A3}0.72  & \cellcolor[HTML]{B6B6B6}0.672          & \cellcolor[HTML]{ABABAB}0.7            & \cellcolor[HTML]{989898}0.747          &  & \cellcolor[HTML]{C1C1C1}0.12  & \cellcolor[HTML]{F4F4F4}0.059 & \cellcolor[HTML]{EAEAEA}0.071 & \cellcolor[HTML]{FDFDFD}0.048 & \cellcolor[HTML]{F5F5F5}0.058 & \cellcolor[HTML]{F9F9F9}0.053 & \cellcolor[HTML]{FFFFFF}\textbf{0.045} \\
2   & \cellcolor[HTML]{D4D4D4}0.599 & \cellcolor[HTML]{F6F6F6}0.516 & \cellcolor[HTML]{FEFEFE}0.497          & \cellcolor[HTML]{EFEFEF}0.534 & \cellcolor[HTML]{FFFFFF}\textbf{0.493} & \cellcolor[HTML]{F5F5F5}0.519          & \cellcolor[HTML]{E8E8E8}0.55           &  & \cellcolor[HTML]{BEBEBE}0.124 & \cellcolor[HTML]{E9E9E9}0.072 & \cellcolor[HTML]{E2E2E2}0.081 & \cellcolor[HTML]{EEEEEE}0.066 & \cellcolor[HTML]{EBEBEB}0.07  & \cellcolor[HTML]{EDEDED}0.067 & \cellcolor[HTML]{F1F1F1}\textbf{0.062} \\
3   & \cellcolor[HTML]{BABABA}0.663 & \cellcolor[HTML]{ECECEC}0.54  & \cellcolor[HTML]{ECECEC}0.54           & \cellcolor[HTML]{E8E8E8}0.551 & \cellcolor[HTML]{F4F4F4}\textbf{0.521} & \cellcolor[HTML]{F2F2F2}0.525          & \cellcolor[HTML]{EDEDED}0.539          &  & \cellcolor[HTML]{B5B5B5}0.135 & \cellcolor[HTML]{DCDCDC}0.088 & \cellcolor[HTML]{D5D5D5}0.096 & \cellcolor[HTML]{DDDDDD}0.086 & \cellcolor[HTML]{DEDEDE}0.085 & \cellcolor[HTML]{E2E2E2}0.081 & \cellcolor[HTML]{E5E5E5}\textbf{0.077} \\
4   & \cellcolor[HTML]{A7A7A7}0.71  & \cellcolor[HTML]{C2C2C2}0.643 & \cellcolor[HTML]{C1C1C1}0.645          & \cellcolor[HTML]{C2C2C2}0.644 & \cellcolor[HTML]{C8C8C8}\textbf{0.629} & \cellcolor[HTML]{C8C8C8}\textbf{0.629} & \cellcolor[HTML]{BEBEBE}0.652          &  & \cellcolor[HTML]{AAAAAA}0.148 & \cellcolor[HTML]{CECECE}0.104 & \cellcolor[HTML]{C4C4C4}0.116 & \cellcolor[HTML]{DDDDDD}0.087 & \cellcolor[HTML]{D1D1D1}0.101 & \cellcolor[HTML]{D6D6D6}0.095 & \cellcolor[HTML]{E2E2E2}\textbf{0.081} \\
5   & \cellcolor[HTML]{A8A8A8}0.707 & \cellcolor[HTML]{CACACA}0.624 & \cellcolor[HTML]{C5C5C5}0.636          & \cellcolor[HTML]{CBCBCB}0.621 & \cellcolor[HTML]{D1D1D1}0.607          & \cellcolor[HTML]{D1D1D1}\textbf{0.606} & \cellcolor[HTML]{CACACA}0.624          &  & \cellcolor[HTML]{ABABAB}0.147 & \cellcolor[HTML]{CECECE}0.105 & \cellcolor[HTML]{C5C5C5}0.115 & \cellcolor[HTML]{D8D8D8}0.093 & \cellcolor[HTML]{D1D1D1}0.101 & \cellcolor[HTML]{D4D4D4}0.097 & \cellcolor[HTML]{DEDEDE}\textbf{0.085} \\
6   & \cellcolor[HTML]{A8A8A8}0.707 & \cellcolor[HTML]{D0D0D0}0.61  & \cellcolor[HTML]{CCCCCC}0.619          & \cellcolor[HTML]{D0D0D0}0.609 & \cellcolor[HTML]{D7D7D7}\textbf{0.592} & \cellcolor[HTML]{D7D7D7}0.593          & \cellcolor[HTML]{D0D0D0}0.61           &  & \cellcolor[HTML]{ACACAC}0.146 & \cellcolor[HTML]{CDCDCD}0.106 & \cellcolor[HTML]{C4C4C4}0.117 & \cellcolor[HTML]{D3D3D3}0.099 & \cellcolor[HTML]{CFCFCF}0.103 & \cellcolor[HTML]{D3D3D3}0.099 & \cellcolor[HTML]{D9D9D9}\textbf{0.091} \\
7   & \cellcolor[HTML]{9A9A9A}0.742 & \cellcolor[HTML]{CACACA}0.624 & \cellcolor[HTML]{C3C3C3}0.642          & \cellcolor[HTML]{CECECE}0.614 & \cellcolor[HTML]{D3D3D3}0.603          & \cellcolor[HTML]{D4D4D4}\textbf{0.6}   & \cellcolor[HTML]{D4D4D4}\textbf{0.6}   &  & \cellcolor[HTML]{ABABAB}0.147 & \cellcolor[HTML]{CCCCCC}0.107 & \cellcolor[HTML]{C3C3C3}0.118 & \cellcolor[HTML]{CCCCCC}0.107 & \cellcolor[HTML]{CFCFCF}0.103 & \cellcolor[HTML]{D1D1D1}0.101 & \cellcolor[HTML]{D3D3D3}\textbf{0.099} \\
8   & \cellcolor[HTML]{9B9B9B}0.738 & \cellcolor[HTML]{CCCCCC}0.618 & \cellcolor[HTML]{C4C4C4}0.638          & \cellcolor[HTML]{D3D3D3}0.602 & \cellcolor[HTML]{D6D6D6}0.594          & \cellcolor[HTML]{D8D8D8}0.59           & \cellcolor[HTML]{D9D9D9}\textbf{0.588} &  & \cellcolor[HTML]{A7A7A7}0.151 & \cellcolor[HTML]{C6C6C6}0.114 & \cellcolor[HTML]{BEBEBE}0.124 & \cellcolor[HTML]{BFBFBF}0.122 & \cellcolor[HTML]{C9C9C9}0.111 & \cellcolor[HTML]{C9C9C9}0.11  & \cellcolor[HTML]{CBCBCB}\textbf{0.108} \\
9   & \cellcolor[HTML]{8D8D8D}0.773 & \cellcolor[HTML]{B8B8B8}0.668 & \cellcolor[HTML]{B1B1B1}0.684          & \cellcolor[HTML]{BEBEBE}0.652 & \cellcolor[HTML]{C2C2C2}0.644          & \cellcolor[HTML]{C3C3C3}\textbf{0.642} & \cellcolor[HTML]{C2C2C2}0.644          &  & \cellcolor[HTML]{A0A0A0}0.16  & \cellcolor[HTML]{BEBEBE}0.124 & \cellcolor[HTML]{B7B7B7}0.132 & \cellcolor[HTML]{BABABA}0.128 & \cellcolor[HTML]{C3C3C3}0.118 & \cellcolor[HTML]{C2C2C2}0.119 & \cellcolor[HTML]{C2C2C2}\textbf{0.119} \\
10  & \cellcolor[HTML]{868686}0.79  & \cellcolor[HTML]{AAAAAA}0.702 & \cellcolor[HTML]{A6A6A6}0.713          & \cellcolor[HTML]{B1B1B1}0.686 & \cellcolor[HTML]{B4B4B4}0.677          & \cellcolor[HTML]{B4B4B4}0.677          & \cellcolor[HTML]{B5B5B5}\textbf{0.675} &  & \cellcolor[HTML]{9B9B9B}0.166 & \cellcolor[HTML]{B9B9B9}0.13  & \cellcolor[HTML]{B3B3B3}0.137 & \cellcolor[HTML]{B1B1B1}0.139 & \cellcolor[HTML]{BEBEBE}0.124 & \cellcolor[HTML]{BDBDBD}0.125 & \cellcolor[HTML]{BABABA}\textbf{0.128} \\
11  & \cellcolor[HTML]{767676}0.83  & \cellcolor[HTML]{9A9A9A}0.74  & \cellcolor[HTML]{9A9A9A}0.742          & \cellcolor[HTML]{A3A3A3}0.718 & \cellcolor[HTML]{A6A6A6}0.713          & \cellcolor[HTML]{A6A6A6}0.712          & \cellcolor[HTML]{A8A8A8}\textbf{0.706} &  & \cellcolor[HTML]{999999}0.168 & \cellcolor[HTML]{B2B2B2}0.138 & \cellcolor[HTML]{ACACAC}0.146 & \cellcolor[HTML]{A5A5A5}0.154 & \cellcolor[HTML]{BABABA}0.128 & \cellcolor[HTML]{B7B7B7}0.132 & \cellcolor[HTML]{B4B4B4}\textbf{0.136} \\
12  & \cellcolor[HTML]{707070}0.843 & \cellcolor[HTML]{979797}0.748 & \cellcolor[HTML]{989898}0.746          & \cellcolor[HTML]{A2A2A2}0.722 & \cellcolor[HTML]{A2A2A2}0.721          & \cellcolor[HTML]{A1A1A1}0.724          & \cellcolor[HTML]{A6A6A6}\textbf{0.713} &  & \cellcolor[HTML]{9C9C9C}0.165 & \cellcolor[HTML]{AEAEAE}0.143 & \cellcolor[HTML]{A5A5A5}0.154 & \cellcolor[HTML]{959595}0.173 & \cellcolor[HTML]{B5B5B5}0.134 & \cellcolor[HTML]{B3B3B3}0.137 & \cellcolor[HTML]{B2B2B2}\textbf{0.138} \\
13  & \cellcolor[HTML]{666666}0.867 & \cellcolor[HTML]{919191}0.762 & \cellcolor[HTML]{949494}0.756          & \cellcolor[HTML]{9F9F9F}0.729 & \cellcolor[HTML]{9E9E9E}0.732          & \cellcolor[HTML]{A0A0A0}0.727          & \cellcolor[HTML]{A3A3A3}\textbf{0.718} &  & \cellcolor[HTML]{9D9D9D}0.163 & \cellcolor[HTML]{ACACAC}0.146 & \cellcolor[HTML]{A3A3A3}0.156 & \cellcolor[HTML]{858585}0.192 & \cellcolor[HTML]{B5B5B5}0.134 & \cellcolor[HTML]{B1B1B1}0.14  & \cellcolor[HTML]{AEAEAE}\textbf{0.143} \\
14  & \cellcolor[HTML]{727272}0.838 & \cellcolor[HTML]{A0A0A0}0.727 & \cellcolor[HTML]{9A9A9A}0.741          & \cellcolor[HTML]{A7A7A7}0.709 & \cellcolor[HTML]{A5A5A5}0.714          & \cellcolor[HTML]{A8A8A8}0.707          & \cellcolor[HTML]{AFAFAF}\textbf{0.691} &  & \cellcolor[HTML]{9D9D9D}0.163 & \cellcolor[HTML]{A5A5A5}0.154 & \cellcolor[HTML]{919191}0.178 & \cellcolor[HTML]{878787}0.19  & \cellcolor[HTML]{B4B4B4}0.136 & \cellcolor[HTML]{ACACAC}0.145 & \cellcolor[HTML]{B2B2B2}\textbf{0.138} \\
15  & \cellcolor[HTML]{7A7A7A}0.82  & \cellcolor[HTML]{A6A6A6}0.713 & \cellcolor[HTML]{A1A1A1}0.724          & \cellcolor[HTML]{ACACAC}0.696 & \cellcolor[HTML]{AFAFAF}0.69           & \cellcolor[HTML]{AFAFAF}0.69           & \cellcolor[HTML]{B8B8B8}\textbf{0.668} &  & \cellcolor[HTML]{A1A1A1}0.159 & \cellcolor[HTML]{A9A9A9}0.149 & \cellcolor[HTML]{898989}0.188 & \cellcolor[HTML]{747474}0.213 & \cellcolor[HTML]{B1B1B1}0.14  & \cellcolor[HTML]{ADADAD}0.144 & \cellcolor[HTML]{B1B1B1}\textbf{0.14}  \\
16  & \cellcolor[HTML]{727272}0.84  & \cellcolor[HTML]{969696}0.751 & \cellcolor[HTML]{949494}0.755          & \cellcolor[HTML]{A2A2A2}0.722 & \cellcolor[HTML]{9F9F9F}0.728          & \cellcolor[HTML]{A2A2A2}0.721          & \cellcolor[HTML]{B4B4B4}\textbf{0.677} &  & \cellcolor[HTML]{A5A5A5}0.154 & \cellcolor[HTML]{A8A8A8}0.15  & \cellcolor[HTML]{8C8C8C}0.184 & \cellcolor[HTML]{666666}0.229 & \cellcolor[HTML]{ADADAD}0.144 & \cellcolor[HTML]{AAAAAA}0.148 & \cellcolor[HTML]{B5B5B5}\textbf{0.135} \\
17  & \cellcolor[HTML]{676767}0.865 & \cellcolor[HTML]{868686}0.79  & \cellcolor[HTML]{8C8C8C}0.775          & \cellcolor[HTML]{8D8D8D}0.772 & \cellcolor[HTML]{939393}0.759          & \cellcolor[HTML]{8D8D8D}0.772          & \cellcolor[HTML]{868686}\textbf{0.791} &  & \cellcolor[HTML]{989898}0.17  & \cellcolor[HTML]{9E9E9E}0.162 & \cellcolor[HTML]{777777}0.209 & \cellcolor[HTML]{6B6B6B}0.224 & \cellcolor[HTML]{ADADAD}0.144 & \cellcolor[HTML]{A4A4A4}0.155 & \cellcolor[HTML]{B9B9B9}\textbf{0.13}  \\ \hline
AVG & \cellcolor[HTML]{8E8E8E}0.771 & \cellcolor[HTML]{B5B5B5}0.674 & \cellcolor[HTML]{B4B4B4}0.678          & \cellcolor[HTML]{BBBBBB}0.661 & \cellcolor[HTML]{BFBFBF}\textbf{0.651} & \cellcolor[HTML]{BEBEBE}0.652          & \cellcolor[HTML]{BEBEBE}0.653          &  & \cellcolor[HTML]{A5A5A5}0.154 & \cellcolor[HTML]{BDBDBD}0.125 & \cellcolor[HTML]{B0B0B0}0.141 & \cellcolor[HTML]{ADADAD}0.144 & \cellcolor[HTML]{C4C4C4}0.117 & \cellcolor[HTML]{C3C3C3}0.118 & \cellcolor[HTML]{C7C7C7}\textbf{0.113} \\ 
\bottomrule
\end{tabular}
\end{table*}

\section{Statistical Metric Learning Benchmark}
ImageNet-21K (IN21K)~\citep{russakovsky2015imagenet}, a superset of ImageNet-1K, comprises 14,197,087 images divided into 21,841 classes. It is organized according to the WordNet~\citep{fellbaum1998wordnet}, providing hierarchical semantics in multiple levels. We propose a novel benchmark, Statistical Metric Learning Benchmark (SMLB), based on IN21K and our proposed metrics to conduct a thorough evaluation of SSL models in the context of metric learning.
This benchmark is meticulously crafted to establish a robust standard for quantifying semantic distance in a large-scale and multiple-grained as well as assessing model generalizability across various contexts.

\subsection{Benchmark Composition}

\paragraph{Data Curation.}
IN21K exhibits a skewed distribution of class representation, with certain classes only containing a minimal number of images, occasionally as low as one. These underrepresented classes fail to offer a dependable intra-class variability. To rectify this, we first selectively curated a subset, hereafter designated as IN20K, which consists of 20,498 classes and 14,191,291 images, by omitting classes comprising less than ten images. Second, we constructed the semantic taxonomy having more than 2 children in accordance with WordNet, where the structure includes 16,632 nodes. Each image is uniformly resized to a resolution of 224x224 pixels and normalized based on the established ImageNet mean and standard deviation parameters.

\begin{figure}[htb]
    \centering
    \includegraphics[width=0.55\linewidth]{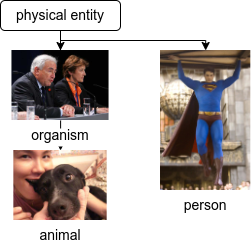}
    \caption{Example of inconsistent tagging in IN21K. Human can be found in  different nodes, e.g., organism, animal, and person.}
    \label{fig:tagging}
\end{figure}

\paragraph{Inconsistency.}
The inherent complexity of IN21K poses challenges in maintaining accurate image classifications, an issue prevalent in large-scale datasets~\citep{DBLP:conf/nips/RidnikBNZ21}. A typical misclassification involves assigning an image having multiple concepts to a closely related class. An instance of such misclassification is depicted in Figure~\ref{fig:tagging}, where identical classes may be represented across different nodes. To avoid instances with the same meaning be regarded as negative samples, we introduce a node-wise taxonomy.

\paragraph{Taxonomic Node.}
In the conceptualization of our benchmark, we utilize a hierarchical semantic taxonomy. A taxonomic node assigns each direct child as one cluster. One cluster consists of all images belonging to the child and its descents. Figure~\ref{fig:evaluation}~\circled{1} graphically illustrates the hierarchical structure of a synset. Notably, he node level indicates the depth within the hierarchical structure; hence, a node positioned at a lower level represents a more abstract, high-level concept.

\subsection{Discriminative Discernment}
This part details our empirical study on the overlap and aSTD across all nodes within the IN20K dataset for six distinct SSL models and one supervised model (DeiT). In the lower-level nodes of the taxonomy, where classes are diverse, distinguishing between them takes few effort. As we ascend to higher-level nodes, the granularity becomes increasingly fine-grained. The results, as summarized in Table~\ref{tab:benchmark}, highlight the effectiveness of SSL methods in semantic learning. It is observed that SD models (DINO and iBOT) suprpass other SSL models and exhibit competitive performance with supervised models (DeiT), particularly within the coarse semantic levels (levels 1-9). 
MIM demonstrates limited effectiveness, accompanied by a notable variance. 
In summation, while a discernible disparity exists between SSL and supervised models, it is not pronounced and predominantly resides within the domain of fine-grained classification.

\subsection{Generalizability}
We elucidate the generalization capabilities of models on downstream tasks through two illustrative scenarios. These scenarios enable the prediction of model performance on distance-based tasks, such as image retrieval and K-Nearest Neighbors (KNN), by examining the overlap on relevant nodes on our benchmark.

\paragraph{Buildings.}
Oxford \& Paris~\citep{DBLP:conf/cvpr/RadenovicITAC18} are benchmarks commonly utilized for the assessment of image retrieval algorithms. To forecast model performance, we identify corresponding nodes via a closely related synset in WordNet, specifically ``building.n.01''. Within this node, we discover 237 children clusters related to the ``building.n.01'' node and compute its overlap.  Then, we compare mean Average Precision (mAP)~\citep{philbin2007object} for the Medium set~\citep{DBLP:conf/cvpr/RadenovicITAC18} with the overlap among these 237 clusters, as presented in Table~\ref{tab:building}.
Despite the differences between Oxford \& Paris and ``building.n.01'' (see comparison in Appendix~\ref{sec:dataset}), the degree of overlap within the "building.n.01" synset serves as a predictive measure for image retrieval performance on the Oxford and Paris datasets. The empirical data suggests that models with lower overlap values tend to exhibit enhanced mAP scores. 

\begin{table}
    \centering
    \caption{Comparison between Image retrieval performance (mAP) for Medium on Oxford \& Paris and overlap on synset ``building.n.01''. The overlap of ``building.n.01'' ranks well the performance on Oxford \& Paris. }
    \label{tab:building}
    \begin{tabular}{lccc}
    \toprule
         &  \multicolumn{2}{c}{mAP $\uparrow$ } &  overlap $\downarrow$ \\
         \cmidrule(lr){2-3}   \cmidrule(lr){4-4}
         & roxford5k& rparis6k& building.n.01\\
\hline
 DINO& 0.3503 (3)& 0.6120 (2)& 0.5486 (2)\\
MAE& 0.1072 (6)& 0.2589 (6)& 0.6962 (6)\\
MSN& \textbf{0.3590 (1)}& 0.5749 (3)& 0.5678 (4)\\
MoCo & 0.2909 (4)& 0.5457 (4)& 0.5618 (3)\\
SiT& 0.2444 (5) & 0.4907 (5) & 0.5922 (5) \\
iBOT& 0.3562 (2)& \textbf{0.6313 (1)}& \textbf{0.5466 (1)}\\
         \bottomrule
    \end{tabular}
\end{table}

\paragraph{Pets and Flowers.} 
We delve into the capability of models to differentiate between highly similar classes -- discriminative discernment. Notably, the Pets~\citep{parkhi2012cats} and Flowers~\citep{DVN_2020_flowers} datasets serve as the benchmarks for this evaluation, with detailed information accessible in Appendix~\ref{sec:dataset}. As delineated in Table~\ref{tab:pets}, a comparative analysis of KNN accuracy and overlap across related synsets demonstrates that overlap is an efficacious indicator of model performance on these nuanced tasks. 
The overlap of a synset indicates the performance of its related downstream tasks.

\begin{table}[tbh]
\caption{Comparison between KNN (k=10) performance (accuracy) on Pets and Flowers and overlap on synsets ``cat.n.01'', ``dog.n.01'', and ``flower.n.01''. Given a synset, the lower overlap between its clusters the higher performance of its related downstream tasks will be.}\label{tab:pets}
\setlength\tabcolsep{2.5pt} 
\begin{tabular}{lcccccc}
\toprule
     & ACC $\uparrow$   & \multicolumn{2}{c}{overlap $\downarrow$} &  & ACC $\uparrow$     & overlap $\downarrow$     \\ 
     \cmidrule(lr){2-2} \cmidrule(lr){3-4}  \cmidrule(lr){6-6} \cmidrule(lr){7-7} 
     & Pets  & cat.n.01     & dog.n.01     &  & Flowers & flower.n.01 \\
 \hline
DINO & \textbf{90.11} & \textbf{0.6648}       & \textbf{0.4946}       &  & \textbf{89.36}   & \textbf{0.5894}      \\
MAE  & 46.96 & 0.8927       & 0.8803       &  & 59.83   & 0.7457      \\
MSN  & 89.83 & 0.6569       & 0.4624       &  & 81.61   & 0.6017      \\
MoCo & 84.76 & 0.6818       & 0.5453       &  & 82.34   & 0.6285      \\
SiT  & 83.83 & 0.7406       & 0.6058       &  & 83.81   & 0.6290      \\
iBOT & 88.91 & 0.6723       & 0.463        &  & 87.56   & 0.5837      \\
\bottomrule
\end{tabular}
\end{table}

\section{Findings}

\begin{figure}[tb]
    \centering
    \includegraphics[width=.8\linewidth]{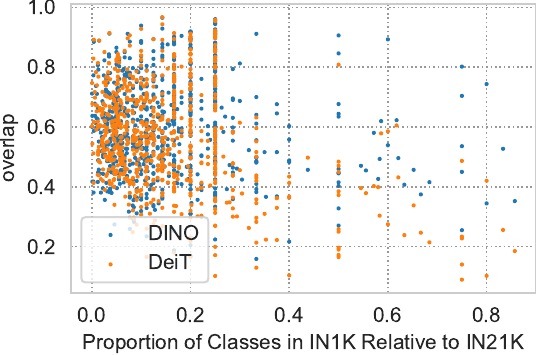}
    \caption{Relationship between class ratio (IN1K vs. IN21K) and overlap for DINO and DeiT. DeiT demonstrates the trend of increased supervision leading to reduced overlap.}
    \label{fig:overlap_label}
\end{figure}

\paragraph{Detrimental Potential of Supervised Learning.}
While supervised learning can ostensibly enhance the acquisition of semantic representations, its success highly depends on the effectiveness of labels.  
As Table~\ref{tab:sup_harm} delineates, the representations of DeiT can be transferred to Pets, yet fail to Flowers.
We observe that there are a small portion of classes related to flowers in IN1K, indicating a limited present of certain classes leads to a skewed knowledge base. Consequently, supervised models suffer from transferring knowledge to a problem with domain shift. Figure~\ref{fig:overlap_label} corroborates this assertion, illustrating the interplay between dataset divergence and class overlap, where X-axis denotes the ratio of the number of classes in IN1K to the number of classes in IN21K for a given synset, Y-axis denotes the respective overlap. Each data points elucidates the degree of separability amidst dataset shift.
Contrastingly, DINO maintains a consistent mean overlap irrespective of the dataset disparity, suggesting an immunity to domain shift. DeiT, however, excels in the presence of small dataset shift (high ratio) but struggles to distinguish between unencountered synsets.

\$ \textit{Is it feasible to retain discriminability amidst supervision?}

\begin{table}[tbh]
\centering
\caption{KNN performance on IN1K does not adequately translate to classify Flowers. Herein, we report KNN precision (k=10) for IN1K, Flowers; the overlap for synsets ``flower.n.01'' and pet (``cat.n.01'' and ``dog.n.01''), respectively.}
\label{tab:sup_harm}
\setlength\tabcolsep{4pt} 
\begin{tabular}{lccccc}
     \toprule
     & IN1K  & Flowers & flower.n.01 &  Pets  &  pet \\
     \midrule
DeiT & 80.82 & 72.69   & 0.6171      &  91.58   & 0.5141 \\
DINO & 76.18 & 89.36   & 0.5894      &  90.11   & 0.5804 \\
\bottomrule
\end{tabular}

\end{table}

\begin{figure}[tb]
    \centering
    \includegraphics[width=0.9\linewidth]{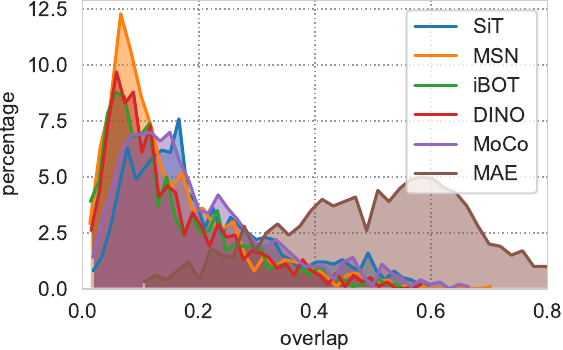}
    \caption{Histogram of overlaps from 1K classes on IN1K for SSL models. Classes that are considered `easy' display lower overlap, while `hard' classes exhibit higher overlap. Though SSL methods are agnostic to the labels, these classes are not learned equally.}
    \label{fig:hist_overlap}
\end{figure}

\begin{figure}[tb]
    \centering
    \includegraphics[width=.8\linewidth]{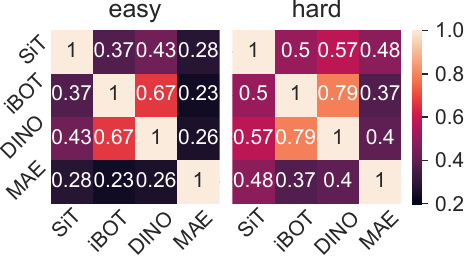}
    \caption{Jaccard similarity coefficient between 100 easy classes (left) and between 100 hard classes (right). The coefficient for random 100 classes is $0.1$. SSL models show agreements with }
    \label{fig:cmp_1k}
\end{figure}

\paragraph{Class Bias of Self-supervised Learning.}
While the class categories within the training dataset are agnostic to SSL models, these models manifest either a propensity or difficulty in internalizing semantic representations from specific class categories. To probe this phenomenon, we assessed the class overlap for each model. The resulting histograms, as depicted in Figure~\ref{fig:hist_overlap}, elucidate a notable pattern: SSL models do not uniformly learn the semantic factors from samples for the classes in the training set. 
In light of this observation, we meticulously compiled two distinct subsets of 100 classes from IN1K for each model; one subset encompassed the `easy' classes characterized by the lowest overlap metrics, while the other comprised the `hard' classes, distinguished by the highest overlap metrics. Subsequently, we applied the Jaccard similarity coefficient to quantify the similarity between two subsets by counting the proportion of shared class commonality. Under the hypothesis of randomness, two arbitrary selections of 100 classes from IN1K would yield a Jaccard coefficient of approximately 0.1. Contrary to this baseline, as delineated in Figure~\ref{fig:cmp_1k}, a pronounced commonality within two subsets was observed among the SSL models.
Notably, MAE demonstrates a lower concordance with other SSL models for the `easy' subset, whereas DINO and iBOT exhibit parallel behaviors for both two subsets. Especially within the `hard' classes, all models indicate high commonality.
Our investigative results cast light on the intrinsic difficulties of SSL models encounter when engaging with the `hard' classes. 

\$ \textit{Which intrinsic limitations within the 'hard' classes present obstacles to the learning process?}

\section{Computational Cost}

We run experiments on a machine with a single RTX 3060, i7-12700, and 32G memory to compare the speed of evaluation with KNN. For a fair comparison, we employ faiss~\citep{johnson2019billion}, a library for efficient similarity search, to run KNN on a single GPU. Table~\ref{tab:runtime} reports the runtime comparison of our protocol with KNN on IN1K for DINO. Each trial repeats 7 times. We can see that our protocol can significantly reduce the runtime by sampling a small proportion of the data without losing performance. The reported runtime excludes the feature extraction.
\begin{table}[htb]
    \centering
\caption{Computational cost comparison with fast KNN on IN1K. Our evaluation can speed up 3.4x without losing performance and 3.9x faster than fast KNN.}
\label{tab:runtime}
\begin{tabular}{lccc}
\toprule
\multicolumn{4}{c}{Statistical Metrics} \\
$n$  & 0.3& 0.5 &   1   \\
t (s$\pm$ms) & 9.66  $\pm$ 77.2  & 16.3  $\pm$ 139  & 33.1s $\pm$ 234  \\
overlap & 0.1344 & 0.1344 & 0.1345   \\
\midrule
\multicolumn{4}{c}{fast KNN} \\
$k$  & 1 & 10 & 20 \\
t (s$\pm$ms)& 38  $\pm$ 245  & 38.5  $\pm$ 255   &39.1  $\pm$ 459  \\
accuracy  & 0.7252 & 0.7490 & 0.7453 \\
\bottomrule
\end{tabular}
\end{table}

\section{Conclusion}

In this study, we introduce the Statistical Metric Learning Benchmark (SMLB) and devise two novel distance-based evaluation metrics -- overlap and aSTD -- to assess the efficacy of semantic representations beyond mere accuracy. Our comprehensive analysis of self-supervised learning (SSL) models, utilizing the SMLB, focuses on their discriminative discernment and generalizability. To the best of our knowledge, SMLB is the first large-scale benchmark that integrates both the diversity and granularity of class categorizations. The key insights garnered from our investigation are threefold: firstly, SSL models demonstrate proficiency in capturing coarse semantic details; secondly, supervised learning approaches hinder the models' ability to adapt to dataset shifts; and thirdly, the 'hard' classes intrinsically challenge the SSL process.

\paragraph{Limitation \& Future Work.} 
While our evaluation methodology is model-agnostic and adaptable to diverse architectures, our current analysis is confined to Vision Transformers (ViTs). Additionally, the 16K nodes within our benchmark encompass instances of mislabeling, a common issue in large-scale datasets. Future endeavors will explore the impact of different architectural choices and refine the evaluation nodes to enhance both efficiency and effectiveness in the evaluation of semantic representation learning.

\newpage
{
    \small
    \bibliographystyle{ieeenat_fullname}
    \bibliography{reference}
}

\clearpage

\section{Dataset}

\begin{table*}[t]
\centering
\caption{Brief summary of IN21K for each level.}\label{tab:21K_info}
\begin{tabular}{l r r rrr}
\toprule
level & classes & samples  & children  &leaf& Examples                                                               \\ \midrule
0     & 21841       & 14197087 & 2         &4& entity, sun, genet, munro, stoker                                          \\
1     & 21837       & 14195721 & 11        &0& physical~entity, abstraction                                            \\
2     & 21837       & 14195721 & 53        &0& object, causal~agent, matter, psychological~feature, thing                 \\
3     & 21837       & 14195721 & 374       &6& whole, person, substance, event, agent                                     \\
4     & 21826       & 14183895 & 884       &149& living~thing, food, artifact, act, infectious~agent                        \\
5     & 21595       & 14058407 & 1849      &459& organism, cell, propulsion, action, activity                               \\
6     & 20971       & 13757813 & 2669      &1124& benthos, heterotroph, animal, plant, jump                                  \\
7     & 19523       & 12958083 & 3250      &1799& hop, check-in, sport, motion                                        \\
8     & 17161       & 11515127 & 3592      &2320& riding, funambulism, rise, contact~sport, outdoor~sport                    \\
9     & 14099       & 9369930  & 3413      &2722& equestrian~sport, climb, acrobatics, track, jumping                        \\
10    & 10639       & 6972298  & 2515      &2711& dressage, rock~climbing, broad~jump, high~jump, bathe                      \\
11    & 7313        & 4906765  & 1799      &2040& curvet, piaffe, fosbury~flop, dead-man's~float, belly~flop                 \\
12    & 4829        & 3367395  & 1138      &1465& ball, professional~baseball, hardball, perfect~game, no-hit~game           \\
13    & 3046        & 2235397  & 802       &902& wolf, dog, bear~cub, soft-finned~fish, spiny-finned~fish                   \\
14    & 1915        & 1426986  & 483       &664& wolf~pup, puppy, ostariophysi, cypriniform~fish, percoid~fish              \\
15    & 1115        & 828458   & 416       &370& loach, cyprinid, sunfish, electric~eel, catostomid                         \\
16    & 636         & 470730   & 188       &344& carp, freshwater~bream, tench, dace, chub                                  \\
17    & 225         & 157373   & 37        &173& domestic~carp, european~bream, common~shiner, buffalo~fish, hog~sucker     \\
18    & 37          & 26490    & 0         &37& leather~carp, mirror~carp, black~buffalo, toy~manchester, sealyham~terrier \\ \bottomrule
\end{tabular}

\end{table*}

\subsection{IN21K details}\label{sec:IN21k}
\begin{figure}[bt]
    \centering
    \includegraphics[width=0.9\linewidth]{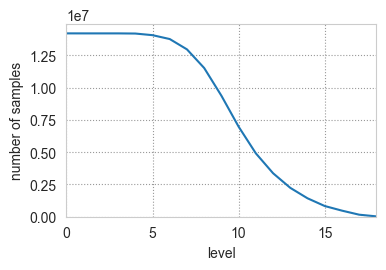}
    \caption{The numbers of classes and samples for each level.}
    \label{fig:21k_level}
\end{figure}

According to the distance to the root node, we assign a level for each node. The node level denotes the depth of the node in the semantic tree. The numbers of classes and samples are shown in Figure~\ref{fig:21k_level}. A detailed summary of each level is in Table~\ref{tab:21K_info}. On the top of the semantic tree, each cluster represents a coarse class consisting of diverse categories. Therefore, the node has more shared features and closer semantic concepts as it increases the level.

\subsection{Dataset Comparison}\label{sec:dataset}

\begin{figure*}[tbh]
    \vspace{-3mm}
    \centering
    \includegraphics[width=0.9\linewidth]{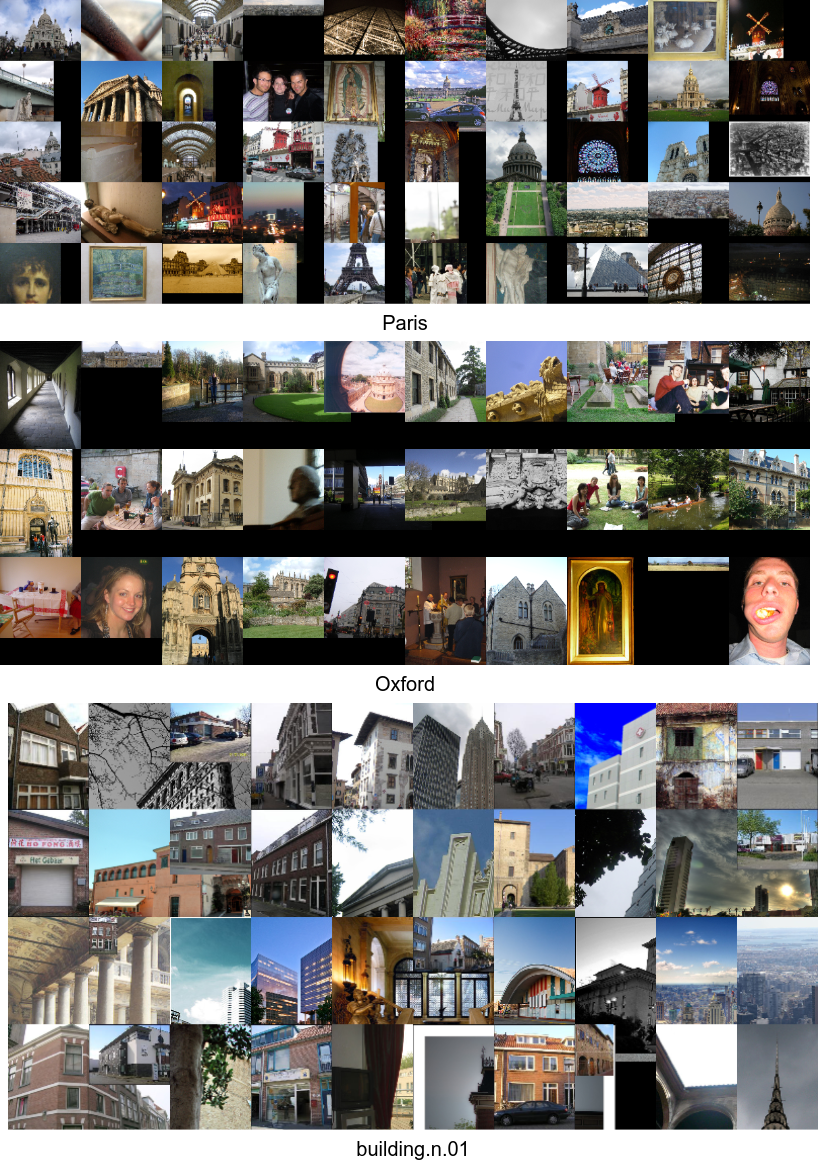}
    \caption{Random samples from Oxford, Paris and ``building.n.01''. One can see notable variability among these datasets: ``building.n.01'' exhibits a broad diversity of structures with minimal noise, like tourists.}
    \label{fig:oxford_paris}
\end{figure*}

\paragraph{Oxford \& Paris.}
The Oxford Building dataset consists of 5,062 images from 11 different landmarks, and Paris dataset contains 6,412 images from Paris landmarks. In IN21K, the closest synset is ``building.n.01'' (``landmark'' does not exist in IN21K). There are 236 children with 194,038 images for ``building.n.01'' from levels 7 to 11. Among them, there are 47 direct children. 
Figure~\ref{fig:oxford_paris} shows samples from Oxford \& Paris as well as the ``building.n.01'' node.
Therefore, the ``building.n.01'' node has far more diversity than Oxford\&Paris. Despite the large disp* between the node and the target datasets, our benchmark can still indicate the performance and show generalizability well.

\paragraph{Dog\&Cat.}
The Pets dataset~\citep{parkhi2012cats} has 12 cat breeds with 2,371 images and 25 dog breeds with 4,978 images. It is notable that there are lots of differences in breeds between Pets~\citep{parkhi2012cats} and ``cat.n.01'' ``dog.n.01''~\citep{russakovsky2015imagenet}. The samples from IN21K contain more irrelative elements, like humans. The categories of breeds in the two sets are also quite different.

\begin{figure}[tbh]
    \centering
    \includegraphics[width=.9\linewidth]{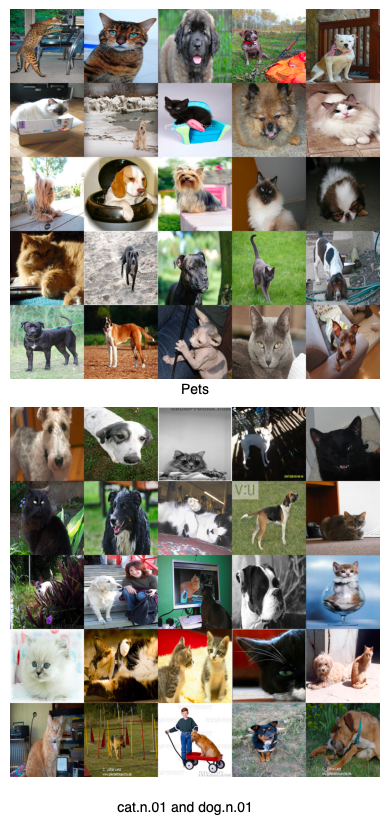}
    \caption{Sample comparison between Pets and the synsets related to dogs and cats in IN21K.}
    \label{fig:enter-label}
\end{figure}

\section{Evaluation Analysis}

\subsection{How important the normalization is?}\label{sec:norm}
Adding BatchNorm without affine transformation after ViT has been proven a helpful trick to improve downstream tasks~\citep{doersch_unsupervised_2015,he_masked_2022}. 
Our observations affirm that the normalization strategy reduces the domain shift. We compared three distinct normalization approaches to ascertain their impact on mitigating domain shift: 1) omission of normalization, 2) application of normalization using the mean and standard deviation of the target dataset, and 3) utilization of the mean and standard deviation from the source dataset. The comparative analysis of overlap for three normalization strategies, as detailed in Table~\ref{tab:cls_norm}, demonstrates that normalizing features using the mean and standard deviation derived from the target dataset yields superior performance across all models. The experimental results suggest that such normalization is effective in domain shift reduction.

\begin{table}[bt]
\centering
\caption{Overlaps on Pets with three normalization strategies: 1) no normalization 2) normalizing with the mean and std from the ``target'' dataset and 3) the ``source'' dataset. MoCo, MAE, and MSN highly rely on normalization.}
\label{tab:cls_norm}
\begin{tabular}{lccc}
\toprule
 Feature Normalization &  None&  Target&  Source\\
 \midrule
DINO & 0.1793 & 0.1544 & 0.1719 \\
MAE & 0.9410 & 0.7199 & 0.7447 \\
MSN & 1.0000 & \textbf{0.1259} & \textbf{0.1375} \\
MoCo & 0.2700 & 0.1955 & 0.2252 \\
iBOT & \textbf{0.1793} & 0.1402 & 0.1552 \\
\bottomrule
\end{tabular}

\end{table}

\subsection{Benchmark for Maximal Overlap}
In previous experiments, we consider the average overlap of one cluster to other clusters. It reveals an overall distance distribution among clusters. It is notable that in the worst case, the overlap between the two most far away clusters is also one of the important criteria for distance metric. Therefore, we further analyze the maximal overlap for each node in Table~\ref{tab:benchmark_max}. There is a similar trend with the average overlap that SSL models perform better on coarse semantic nodes. 

\begin{table}[]
\centering
\caption{Comparative analysis of maximal overlap for SSL models and DeiT on 16K synsets grouped by node levels. `AVG' denotes the average metrics across all nodes. DeiT surpasses others from levels 7-16.}\label{tab:benchmark_max}
\setlength\tabcolsep{3pt} 
\begin{tabular}{lccccccc@{}}
\toprule
 & MAE                           & MoCo                          & SiT                           & MSN                                    & DINO                                   & iBOT                                   & DeiT                                   \\
\midrule
1     & \cellcolor[HTML]{E9E9E9}0.942 & \cellcolor[HTML]{D1D1D1}0.909 & \cellcolor[HTML]{DBDBDB}0.923 & \cellcolor[HTML]{D7D7D7}0.917          & \cellcolor[HTML]{C8C8C8}\textbf{0.896} & \cellcolor[HTML]{CBCBCB}0.9            & \cellcolor[HTML]{E7E7E7}0.94           \\
2     & \cellcolor[HTML]{A7A7A7}0.849 & \cellcolor[HTML]{7B7B7B}0.786 & \cellcolor[HTML]{6B6B6B}0.764 & \cellcolor[HTML]{666666}\textbf{0.756} & \cellcolor[HTML]{727272}0.774          & \cellcolor[HTML]{7D7D7D}0.789          & \cellcolor[HTML]{7D7D7D}0.789          \\
3     & \cellcolor[HTML]{B4B4B4}0.868 & \cellcolor[HTML]{7B7B7B}0.786 & \cellcolor[HTML]{7F7F7F}0.792 & \cellcolor[HTML]{929292}0.819          & \cellcolor[HTML]{767676}0.78           & \cellcolor[HTML]{777777}0.781          & \cellcolor[HTML]{767676}\textbf{0.779} \\
4     & \cellcolor[HTML]{CECECE}0.904 & \cellcolor[HTML]{B0B0B0}0.862 & \cellcolor[HTML]{B3B3B3}0.866 & \cellcolor[HTML]{ADADAD}0.857          & \cellcolor[HTML]{A8A8A8}\textbf{0.85}  & \cellcolor[HTML]{ABABAB}0.854          & \cellcolor[HTML]{B0B0B0}0.862          \\
5     & \cellcolor[HTML]{CACACA}0.898 & \cellcolor[HTML]{A8A8A8}0.85  & \cellcolor[HTML]{ADADAD}0.857 & \cellcolor[HTML]{A4A4A4}0.845          & \cellcolor[HTML]{A0A0A0}0.839          & \cellcolor[HTML]{9F9F9F}\textbf{0.838} & \cellcolor[HTML]{A4A4A4}0.844          \\
6     & \cellcolor[HTML]{BDBDBD}0.88  & \cellcolor[HTML]{959595}0.824 & \cellcolor[HTML]{9C9C9C}0.833 & \cellcolor[HTML]{969696}0.825          & \cellcolor[HTML]{8F8F8F}\textbf{0.815} & \cellcolor[HTML]{8F8F8F}\textbf{0.815} & \cellcolor[HTML]{959595}0.823          \\
7     & \cellcolor[HTML]{CACACA}0.898 & \cellcolor[HTML]{9A9A9A}0.83  & \cellcolor[HTML]{A0A0A0}0.839 & \cellcolor[HTML]{909090}0.816          & \cellcolor[HTML]{8E8E8E}0.814          & \cellcolor[HTML]{8F8F8F}0.815          & \cellcolor[HTML]{8B8B8B}\textbf{0.809} \\
8     & \cellcolor[HTML]{C0C0C0}0.885 & \cellcolor[HTML]{8E8E8E}0.813 & \cellcolor[HTML]{969696}0.825 & \cellcolor[HTML]{858585}0.801          & \cellcolor[HTML]{828282}0.797          & \cellcolor[HTML]{808080}0.794          & \cellcolor[HTML]{7D7D7D}\textbf{0.79}  \\
9     & \cellcolor[HTML]{CFCFCF}0.905 & \cellcolor[HTML]{A6A6A6}0.847 & \cellcolor[HTML]{ADADAD}0.857 & \cellcolor[HTML]{9C9C9C}0.833          & \cellcolor[HTML]{9A9A9A}0.831          & \cellcolor[HTML]{9A9A9A}0.83           & \cellcolor[HTML]{989898}\textbf{0.828} \\
10    & \cellcolor[HTML]{D8D8D8}0.918 & \cellcolor[HTML]{B8B8B8}0.873 & \cellcolor[HTML]{BCBCBC}0.878 & \cellcolor[HTML]{B0B0B0}0.861          & \cellcolor[HTML]{AEAEAE}0.859          & \cellcolor[HTML]{ADADAD}0.858          & \cellcolor[HTML]{ABABAB}\textbf{0.854} \\
11    & \cellcolor[HTML]{E4E4E4}0.936 & \cellcolor[HTML]{BEBEBE}0.881 & \cellcolor[HTML]{BEBEBE}0.881 & \cellcolor[HTML]{B4B4B4}0.868          & \cellcolor[HTML]{B3B3B3}0.866          & \cellcolor[HTML]{B2B2B2}0.865          & \cellcolor[HTML]{B0B0B0}\textbf{0.862} \\
12    & \cellcolor[HTML]{E5E5E5}0.937 & \cellcolor[HTML]{BFBFBF}0.883 & \cellcolor[HTML]{BCBCBC}0.878 & \cellcolor[HTML]{B4B4B4}0.868          & \cellcolor[HTML]{B4B4B4}0.867          & \cellcolor[HTML]{B4B4B4}0.868          & \cellcolor[HTML]{B2B2B2}\textbf{0.865} \\
13    & \cellcolor[HTML]{EAEAEA}0.944 & \cellcolor[HTML]{C1C1C1}0.886 & \cellcolor[HTML]{BFBFBF}0.883 & \cellcolor[HTML]{B5B5B5}0.869          & \cellcolor[HTML]{B9B9B9}0.874          & \cellcolor[HTML]{B4B4B4}0.867          & \cellcolor[HTML]{AEAEAE}\textbf{0.859} \\
14    & \cellcolor[HTML]{EBEBEB}0.946 & \cellcolor[HTML]{C3C3C3}0.888 & \cellcolor[HTML]{C8C8C8}0.895 & \cellcolor[HTML]{B3B3B3}0.866          & \cellcolor[HTML]{BBBBBB}0.877          & \cellcolor[HTML]{B9B9B9}0.875          & \cellcolor[HTML]{ADADAD}\textbf{0.858} \\
15    & \cellcolor[HTML]{DDDDDD}0.926 & \cellcolor[HTML]{B0B0B0}0.861 & \cellcolor[HTML]{B6B6B6}0.87  & \cellcolor[HTML]{ABABAB}0.854          & \cellcolor[HTML]{A5A5A5}0.846          & \cellcolor[HTML]{A4A4A4}0.845          & \cellcolor[HTML]{9C9C9C}\textbf{0.834} \\
16    & \cellcolor[HTML]{DEDEDE}0.927 & \cellcolor[HTML]{C0C0C0}0.884 & \cellcolor[HTML]{C2C2C2}0.887 & \cellcolor[HTML]{AAAAAA}0.853          & \cellcolor[HTML]{B3B3B3}0.866          & \cellcolor[HTML]{B4B4B4}0.868          & \cellcolor[HTML]{898989}\textbf{0.807} \\
17    & \cellcolor[HTML]{FFFFFF}0.973 & \cellcolor[HTML]{E2E2E2}0.932 & \cellcolor[HTML]{E0E0E0}0.93  & \cellcolor[HTML]{E7E7E7}0.94           & \cellcolor[HTML]{E0E0E0}\textbf{0.93}  & \cellcolor[HTML]{E3E3E3}0.934          & \cellcolor[HTML]{E3E3E3}0.934          \\

AVG   & \cellcolor[HTML]{D5D5D5}0.914 & \cellcolor[HTML]{AEAEAE}0.859 & \cellcolor[HTML]{B0B0B0}0.862 & \cellcolor[HTML]{A8A8A8}0.85           & \cellcolor[HTML]{A5A5A5}0.846          & \cellcolor[HTML]{A6A6A6}0.847          & \cellcolor[HTML]{A3A3A3}0.843         \\
\bottomrule
\end{tabular}
\end{table}

\subsection{Object-wise Similarity.}

In this part, we extend our evaluation of local features to study their separability and consistency.
Similarly, to study the local features within an image ${\vx}$, we define the \emph{intra-object} pair as the pair of image patches from the same object, and \emph{inter-object} pair as the pair of image patches from different objects. For a given set of image patches ${\vx_p}$ and their corresponding masks ${\vm_p}$ that identify the object's location, we can determine if patches originate from the same or different objects. Thus, we can define the intra- and inter-object similarity as:
\begin{equation}
\begin{aligned}
    s^+_\mathtt{OBJ}(\vx_i,\vx_j) &= \mathit{sim}(\vz_i,\vz_j), \text{if}\ \vm_i = \vm_j, \\
    s^-_\mathtt{OBJ}(\vx_i,\vx_j) &= \mathit{sim}(\vz_i,\vz_j), \text{if}\ \vm_i \neq \vm_j.
\end{aligned}
\label{eq:object_similarity}
\end{equation}

Our protocol provides a unified method for assessing the representations for both global and local semantics. For global semantics, we employ class labels to define positive pairs. In contrast, we utilize object segmentation to define positive pairs for local semantics. We collect the local features for each object over the whole dataset and compute the statistical metrics for two models: MAE~\cite{he_masked_2022} and iBOT~\cite{zhou_ibot_2022}. We considered three normalization methods for the local features using the mean and std along the sample dimension (0), sequence dimension (1), and both (2).
To compare the quality of dense features for SSL models, we conduct experiments on ImageNetS, a variant of ImageNet with semantic segmentation~\citep{zhai_large-scale_2020}. This dataset contains segmentation labels of the dominant object for images. 919 subset is the largest split with 12,419 images divided into 919 categories.  
Table~\ref{tab:hist_pat} demonstrates overlap and aSTD on ImageNet-S 919 subset for MAE and iBOT with three normalization types. 
Though MAE has limited performance on global features, its local features have higher quality in indicating semantic distance than iBOT.

\begin{table}[tb]
\caption{Object-wise statistical measurement on ImageNet-S 919 subset. ``fn\_type'' denotes feature normalization type of the direction of normalization: 0 for the samples, 1 for the patches within an image, 2 for both.}
\label{tab:hist_pat}
\begin{tabular}{@{}lcccccc@{}}
\toprule
         & \multicolumn{3}{c}{aSTD} & \multicolumn{3}{c}{overlap} \\
\cmidrule(lr){2-4} \cmidrule(lr){5-7}
fn\_type & 0      & 1      & 2      & 0       & 1       & 2       \\
MAE      & \textbf{0.060} & 0.088 & 0.087 & 0.630 & 0.763 & \textbf{0.576} \\
iBOT     & \textbf{0.074} & 0.117 & 0.116 & 0.649 & 0.737 & \textbf{0.604}\\
\bottomrule
\end{tabular}
\end{table}

\section{Efficiency}

\paragraph{Sampling.}
Our proposed metrics are based on statistical distributions. Therefore, the quality of estimation is vital for the final result, which is affected by the number of samples. We study the estimation error of sampling a part of the samples in the data by comparing the metrics in the full dataset within a proportion of the dataset. The experiments are finished on a machine with a single RTX 3060, i7-12700, and 32G memory.
Table~\ref{tab:sampling} reports the spending time, KL, overlap, and aSTD for randomly picking $n$ percent of samples. We can see that sampling 30\% percent of data saves 91\% evaluation time, maintaining similar results.
\begin{table}[htb]
    \centering
    \caption{Estimation error for $n$ percent of the data. }
    \label{tab:sampling}
\begin{tabular}{lcccc}
\toprule
      $n$ &     time (s) &   overlap &  inter-aSTD & intra-aSTD \\
\midrule
  0.01 &    7.3486 &  0.1450 &  0.0553 &  0.2518 \\
  0.1 &  228.2575 &  0.1284 &  0.0554 &  0.2202 \\
  0.3 &   1010.3546 &  0.1261 &  0.0554 &  0.2179 \\
  0.5 &   2818.6852 &  0.1258 &  0.0554 &  0.2175 \\
  1.0 &  11161.6814 &  0.1255 &  0.0554 &  0.2171 \\

\bottomrule
\end{tabular}
\end{table}

\begin{figure}[hb]
    \centering
    \includegraphics[width=.9\linewidth]{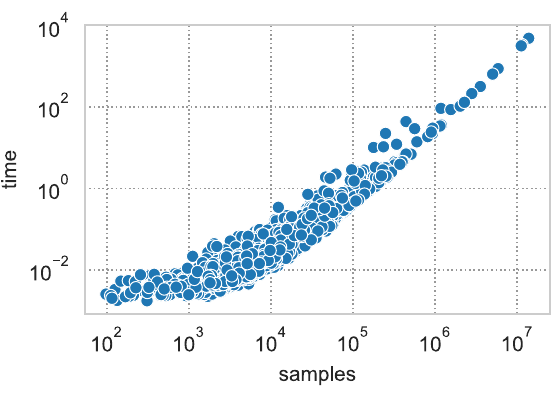}
    \caption{Computation cost of nodes. Each point demonstrates the evaluation time of a node without feature extraction and its sample number.}
    \label{fig:runtime}
\end{figure}

\paragraph{Runtime.}
Our evaluation is effective in assessing representations on our large-scale benchmark. The whole evaluation process consists of two parts: feature extraction and computing overlaps. We evaluated each model on a machine with 4 RTX 2080 Ti GPUs, AMD EPYC 7502 (32 cores), and 256G memory. The runtime for feature extraction using 4 GPUs is 11 hours. Then, we compute the overlaps for all synset nodes with a single GPU. Figure~\ref{fig:runtime} shows the runtime of each node and its number of samples. The total computation time is 14,747 seconds. 

\end{document}